\documentclass[runningheads]{llncs}

\usepackage{amsmath,amssymb}
\usepackage{bm}

\usepackage{booktabs}
\usepackage{array}
\usepackage{multirow}
\usepackage{tabularx}

\usepackage{graphicx}
\usepackage{float}
\usepackage{tikz}
\usetikzlibrary{shapes.geometric, arrows.meta, positioning, fit, backgrounds}

\usepackage[ruled,linesnumbered]{algorithm2e}
\SetKwComment{Comment}{// }{}

\usepackage[colorlinks=true,linkcolor=black,citecolor=black,urlcolor=blue]{hyperref}

\title{Making Failure Safe: A Constrained, Verifiable Agent Framework for Open-Web Data Collection}
\titlerunning{A Constrained, Verifiable Agent Framework for Open-Web Data Collection}

\author{Bo Chen}
\authorrunning{B. Chen}
\institute{Institute of Computing Technology, Chinese Academy of Sciences, Beijing, China\\
\email{chenbo01@ict.ac.cn}}

\begin{document}

\maketitle

\begin{abstract}
LLMs and agents can generate web scrapers from natural-language requirements, but direct generation remains unreliable because of dependency errors, broken selectors, schema mismatches, and heterogeneous page structures. We propose a constrained, verifiable agent framework that shifts LLM output from free-form code to typed JSON collector configurations, combining a six-type collector taxonomy, template and utility-function constraints, static Airflow DAG execution, rule-based quality checking, and structured feedback correction. Experiments on 138 tasks show that the taxonomy supports description-based requirement typing, while confirming that stable instantiation requires completing source, field, and execution constraints beyond the initial description. On 80 independently source-verified tasks, the framework runs with zero execution-stage LLM tokens and the lowest average wall-clock time, trading moderate one-shot quality for a reusable, deterministic, and verifiable execution path suited to repeated scheduled collection. These results position the framework as a reusable, low-cost, and verifiable execution path for repeated open-web data collection.

\end{abstract}


\section{Introduction}

Open-web data---publicly accessible information from news, government notices, e-commerce, and academic publications---drives growing demand for automated collection at scale. Traditional pipelines depend on manual effort: requirement analysis, site-structure inspection, scraper coding, and validation, suffering from long development cycles and low reusability.

Recent LLM and agent advances offer a new path: understanding natural-language requirements, generating scraper code, and executing validation tasks. However, direct application proves unreliable. The core tension is between the \textbf{structural brittleness} of heterogeneous web pages and the \textbf{stochastic inconsistency} of LLM code generation. Without collector-type constraints, agents deviate in task decomposition; LLM-generated scrapers may contain dependency errors, broken selectors, and inconsistent output schemas.

To address this, we argue that the key is not to pursue perfect agent reasoning, but to make the collection pipeline \textbf{verifiable}---so that every stage's inputs and outputs can be checked, measured, and audited. We achieve this through three pillars: (1) typed modeling of collection tasks with explicit functional boundaries; (2) template and utility-function constraints that shift the agent from free-form code generation to configuration generation; and (3) small-sample validation and quality feedback before scale execution, forming a ``generate--execute--check--fix'' closed loop.

Our contributions are:
\begin{enumerate}
\item A \textbf{six-type collector taxonomy} (search, list, detail, API, interactive, file) with explicit functional boundaries and composition rules.
\item A \textbf{constrained agent framework} organizing requirement understanding, instantiation, validation, quality checking, and feedback correction into a verifiable closed loop, generating JSON configurations under template, slot, and Schema constraints---not free-form code.
\item \textbf{Multi-level experimental validation} covering requirement-type classification, collector instantiation, quality--cost comparison against runtime LLM baselines on 80 verified tasks, feedback correction, and Airflow compatibility testing.
\end{enumerate}

The research is organized around three questions: \textbf{RQ1: How to model collection requirements as typed tasks}; \textbf{RQ2: How to instantiate collectors under constraints for validatable, reusable, schedulable configurations}; \textbf{RQ3: How to intercept low-quality configurations before scale execution via validation and feedback}. These are addressed through the framework design in Sections~3--5 and answered in Section~7. The framework embodies a design philosophy: \textbf{controlled, explicit failure is preferable to uncontrolled, silent success}---through type constraints, Schema validation, and rule-based quality gating, failures become auditable, traceable signals.

\section{Related Work}

\subsection{Open-Web Data Collection}

Traditional web data collection builds on Scrapy, BeautifulSoup, Selenium, and Playwright, orchestrated via workflow engines. The central challenge is website heterogeneity and volatility, which forces heavy manual effort~\cite{kushmerick1997,ferrara2014}. A recent review of 91 papers~\cite{landeta2026} notes that 84\% appeared in 2024--2025, trending from rule-based to semantic and agent-based methods. Tools such as ScrapeGraphAI~\cite{scrapegraphai} and Crawl4AI~\cite{crawl4ai} leverage LLMs to generate scraper logic but introduce execution unpredictability. MacroBench~\cite{macrobench2025} tests LLM web automation on 681 tasks: simple tasks achieve 91.7\% success but complex multi-step workflows score 0.0\%. Webscraper~\cite{webscraper2026} uses multimodal LLMs with five-stage prompting for index-content architectures. AutoScraper~\cite{autoscraper2024} proposes progressive understanding with an executability metric. These works demonstrate rapid evolution toward agent-based scraping, but unconstrained code generation remains unreliable in complex scenarios.

\subsection{LLMs, Code Generation, and Constraints}

LLMs show strong code generation capabilities~\cite{chen2021,nijkamp2022,roziere2023}, but directly generated code suffers from API errors, dependency inconsistencies, and format mismatches. Reflexion~\cite{shinn2023} and Self-Refine~\cite{madaan2023} use language-feedback-based reflection, yet Yu et al.~\cite{yu2024} show that LLMs often fail to identify errors in their own code, revealing clear boundaries to self-correction. In constrained generation, PARSE~\cite{parse2025} optimizes JSON schemas for LLM extraction with static and LLM guardrails, achieving up to 64.7\% accuracy improvement. llm-scraper~\cite{llmscraper2024} uses Zod schemas to constrain Playwright DOM selector code. Both focus on single-layer constraints. Our framework introduces a four-tier nested constraint hierarchy (type $\rightarrow$ template $\rightarrow$ utility $\rightarrow$ validation) with collector-type and utility-function reuse constraints beyond schema-level restrictions.

\subsection{Agents and Web Automation}

The ReAct~\cite{yao2022} paradigm interleaves reasoning and action for web tasks, evaluated by benchmarks such as WebVoyager~\cite{he2024} and WebArena~\cite{zhou2024}. In collection practice, BardeenAgent~\cite{bardeenagent2025} proposes Record-then-Replay: the LLM learns extraction patterns during recording, then a deterministic program replays without LLM calls, achieving 66.2\% recall on WebLists at one-third the per-row cost. CyberScribe~\cite{cyberscribe2025} combines LLMs, vision transformers, and reinforcement learning for dynamic multi-language collection. AutoData~\cite{autodata2025} uses eight specialized agents with OHCache to reduce inter-agent token overhead, but its Engineering Agent still generates free-form code without a collector type taxonomy or schedule-native quality gating. Unlike ReAct-style end-to-end browsing and BardeenAgent's operation-sequence decoupling, we invoke the agent only during requirement understanding and collector instantiation, while execution runs through a static DAG, decoupling JSON configurations rather than browser operation sequences.

\subsection{Constrained Generation and Quality Feedback}

Data quality research emphasizes accuracy, completeness, consistency, and fitness for use~\cite{wang1996,pipino2002}. In LLM-involved pipelines, quality issues also include selector errors, field-mapping errors, and execution anomalies. The AI Committee~\cite{aicommittee2025} uses four specialized agents for collaborative validation, achieving 78.7\% completeness and up to 100\% precision. Berkane et al.~\cite{berkane2025} constrain LLM extraction via Pydantic schemas with source-grounding verification and LLM-based anomaly detection, achieving +74.3 and +38.5 percentage point F1 improvements, though quality control still relies on LLM subjective judgment.

In feedback correction, methods such as Reflexion~\cite{shinn2023} and Self-Refine~\cite{madaan2023} use intrinsic self-correction, while Agent-R~\cite{agentr2025} uses Monte Carlo Tree Search. However, LLMs often fail at self-error-detection~\cite{yu2024}. Our approach differs fundamentally: we employ rule-based quality checks as an external evaluator, convert quality reports into structured feedback constraints (not natural-language reflections), use a failure blacklist for monotonic pruning, and have the agent regenerate under these constraints. This ``external rule evaluation $\rightarrow$ structured constraints $\rightarrow$ blacklist pruning $\rightarrow$ constrained regeneration'' paradigm also introduces original-page-archive comparison to distinguish \texttt{source\_defect} from \texttt{parse\_error}.

\section{Problem Definition}

Given a user's initial task description $d$ (natural language, possibly with partial structured information), the system must complete requirement understanding, collector type selection, instantiation, validation execution, scheduling, and quality feedback.

Let the full task representation be $R = (d, s, x, f, c, o)$, where $s$ is the target source identifier, $x$ is source context from proactive probing, $f$ is the field set, $c$ denotes collection constraints, and $o$ is the output format. Let $\mathcal{M} = (C, P, V)$ be the constrained configuration tuple, where $C$ is the collector type or composition, $P$ contains runtime parameters, and $V$ specifies validation thresholds. The system output is $O = (\mathcal{M}, E, D, Q)$, where $E$ represents Airflow DAG execution, $D$ is collected data, and $Q$ is the quality feedback report.

Unlike unconstrained code generation, we formulate instantiation as template-constrained slot assignment. Let $\mathcal{T}$ be the template space and $\Theta_k$ the slot space of template $\mathcal{T}_k$. The system must find $\theta_k \in \Theta_k$ such that $\mathcal{M} = \mathrm{Instantiate}(\mathcal{T}_k, \theta_k)$ and $\mathrm{Check}(\mathcal{M}) = \mathrm{True}$.

\section{Collector Type Taxonomy}

The taxonomy follows five principles: entry-point orientation, data-shape orientation, single responsibility, explicit boundaries, and scheduler composability. Common compositions include Search+Detail (keyword news), List+Detail+File (announcements), and Interactive+List+Detail (dynamic forms).

\smallskip
\noindent\textbf{Search.} Discovers candidate URLs by keyword from search entry points. Takes keywords and a time range, outputs result links and titles. Boundary: no content extraction. Typical scenario: site search.

\noindent\textbf{List.} Traverses paginated list pages to discover detail-page links. Takes a column URL and page range, outputs list items and detail links. Boundary: no deep content extraction. Typical scenario: news/announcement lists.

\noindent\textbf{Detail.} Extracts structured fields (title, body, author, timestamp) from a single detail page. Takes a detail-page URL. Boundary: no large-scale URL discovery. Typical scenario: news/announcement detail pages.

\noindent\textbf{API.} Calls public APIs and maps response fields. Takes an endpoint URL and request parameters, outputs structured records. Boundary: no browser interaction. Typical scenario: JSON REST APIs.

\noindent\textbf{Interactive.} Executes clicks, text input, and page actions on dynamically rendered pages. Takes a page URL and action sequence, outputs rendered data. Boundary: no bypassing authentication or CAPTCHAs. Typical scenario: dynamically loaded pages.

\noindent\textbf{File.} Downloads and parses publicly accessible files (PDF, Excel). Takes a file URL, outputs file text and metadata. Boundary: no semantic judgment. Typical scenario: PDF/Excel attachments.
\smallskip

\section{Constrained Agent Collection Framework}

\subsection{Overview}

Our framework transforms unconstrained agent collection into a constrained closed loop through four constraint categories: (1) \textbf{task-type constraint}---the requirement understanding agent determines the collector type, preventing configuration generation without a clear collection pattern; (2) \textbf{configuration-structure constraint}---the instantiation agent is constrained by template slots and JSON Schema; (3) \textbf{tool-reuse constraint}---the framework reuses pre-built utility functions through templates rather than generating request, parse, or cleaning code directly; (4) \textbf{execution-feedback constraint}---every new configuration must pass small-sample validation and quality assessment before scale execution, with feedback triggering correction on failure.

\begin{figure}[H]
\centering
\begin{tikzpicture}[
    node distance=0.24cm,
    data/.style={
        ellipse, draw=black!55, line width=0.4pt,
        minimum width=4.4cm, minimum height=0.42cm,
        align=center, font=\fontsize{6.2}{7.2}\selectfont, inner sep=1.5pt, fill=gray!3,
    },
    proc/.style={
        rectangle, rounded corners=1pt, draw=black!55, line width=0.4pt,
        minimum width=4.4cm, minimum height=0.32cm,
        align=center, font=\fontsize{6.2}{7.2}\selectfont, inner sep=1.5pt,
    },
    agent/.style={proc, fill=blue!3},
    exec/.style={proc, fill=green!3},
    fix/.style={proc, fill=red!4, minimum width=2.0cm},
    decision/.style={
        diamond, draw=black!55, line width=0.4pt, aspect=2.8,
        minimum width=1.0cm, minimum height=0.42cm,
        align=center, font=\fontsize{6.2}{7.2}\selectfont, inner sep=0pt,
    },
    arr/.style={->, >=Stealth, line width=0.45pt, black!70},
    farr/.style={->, >=Stealth, line width=0.5pt, red!60, densely dashed},
]
\node[data] (input) {User initial task description $d$};
\node[agent, below=of input] (classifier) {Requirement Understanding Agent};
\node[data, below=of classifier] (structured) {Full task representation $R$};
\node[agent, below=of structured] (template) {Collector Instantiation Agent};
\node[data, below=of template] (config) {Config $\mathcal{M}$ (contains $C,P,V$)};
\node[exec, below=of config] (execute) {Airflow Execution (validation stage)};
\node[exec, below=of execute] (qa) {Quality Check};
\node[decision, below=0.35cm of qa] (dec) {Pass?};
\node[exec, below=0.35cm of dec] (scale) {Airflow Execution (scale stage)};
\node[data, below=of scale] (result) {Results $D$ and Quality Report $Q$};
\node[fix, left=0.7cm of dec] (feedback) {Feedback Correction};

\draw[arr] (input) -- (classifier);
\draw[arr] (classifier) -- (structured);
\draw[arr] (structured) -- (template);
\draw[arr] (template) -- (config);
\draw[arr] (config) -- (execute);
\draw[arr] (execute) -- (qa);
\draw[arr] (qa) -- (dec);
\draw[arr] (dec) -- node[right, font=\scriptsize] {Yes} (scale);
\draw[arr] (scale) -- (result);
\draw[arr] (dec) -- node[above, font=\scriptsize] {No} (feedback);
\draw[farr] (feedback.north) |- (template.west)
    node[pos=0.72, left, font=\scriptsize, text=red!60] {Feedback constraints $\Gamma_t$};
\end{tikzpicture}
\caption{Constrained agent collection framework architecture}
\label{fig:architecture}
\end{figure}
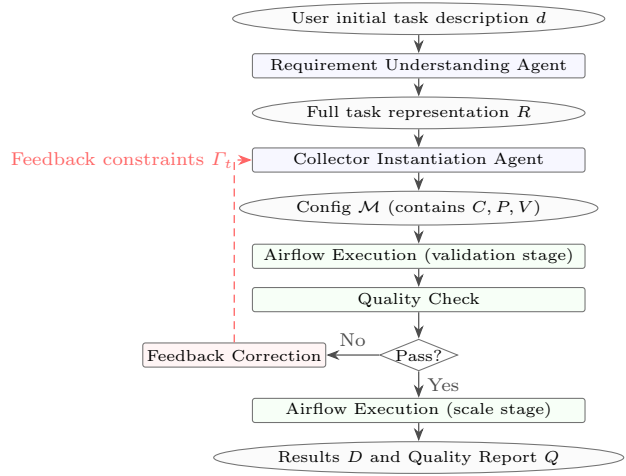

The framework consists of five modules: requirement understanding agent, collector instantiation agent, Airflow validation/scale execution, quality check, and feedback correction (Fig.~\ref{fig:architecture}).

\subsection{Requirement Understanding Agent}

The requirement understanding agent takes the user's initial description $d$ and completes $R=(d,s,x,f,c,o)$. For explicitly provided components, the agent writes them directly into slots; for missing components, it probes the target site (homepage, navigation, robots/sitemap, API docs, sample pages) to extract page titles, column structures, candidate URLs, DOM summaries, pagination clues, and sample records, compressed into source context $x$.

Let $\mathrm{Probe}$ denote the probing function:
\begin{equation}
x = \mathrm{Probe}(d), \quad (s,f,c,o,\kappa) = \Phi(d,x), \quad R = (d,s,x,f,c,o),
\end{equation}
where $\kappa$ denotes the collector type or composition, and $\Phi$ decomposes into per-slot sub-mappings. The output is constrained to a fixed Schema with typed slots for task type, collector types, target sources, source context, required fields, constraints, output schema, validation config, and confidence.

\subsection{Collector Instantiation Agent}

The agent uses ``reference coding + template slot-filling'': templates hard-code Airflow structure, utility-function skeletons, error handling, and persistence; the agent fills in URLs, selectors, API parameters, field mappings, and cleaning rules. In the first round, $R$ drives template retrieval and slot filling; in correction rounds, the quality report $Q$ is converted into feedback constraints fed alongside $R$. The prompt context assembles: (1) System Prompt with collector type specs, Schema constraints, and compliance rules; (2) Few-Shot examples retrieved by type and task similarity; (3) the full task representation; (4) slot-filling instructions.

Let $\mathcal{U}$ be the utility function set and $\mathcal{P}$ the prompt context space. The prompt constructor $\mathrm{Prompt}: \mathcal{T} \times 2^{\mathcal{U}} \rightarrow \mathcal{P}$ yields $p_k = \mathrm{Prompt}(\mathcal{T}_k, \mathcal{U})$. The agent generates slot assignments under joint constraints:
\begin{equation}
\theta_k^* = \mathrm{LLM}(R, p_k), \quad \theta_k^* \in \Theta_k, \quad
\mathcal{M} = \mathrm{Instantiate}(\mathcal{T}_k, \theta_k^*), \quad \mathrm{Check}(\mathcal{M}) = \mathrm{True}.
\end{equation}
The agent outputs a Schema-compliant JSON configuration $\mathcal{M}$ (not executable code), which the generic collector operator reads and executes interpretatively.

\subsection{Airflow Validation and Scale Execution}

Validation and scale runs share the same core logic: a generic collector operator reads the JSON configuration and invokes pre-built utility functions. Validation runs use small-sample parameters (limited pages, output rows, forced original-page archiving) to shift errors to a low-cost pre-production stage; scale runs use production parameters. A static execution DAG with dynamic configuration interpretation prevents LLM-generated code from breaking Airflow's Scheduler parsing.

\subsection{Quality Check Module}

This module evaluates validation-run results using rule-based, deterministic checks (no LLM required). It scores five dimensions and computes a composite score:
\begin{equation}
S = w_s S_s + w_c S_c + w_r S_r + w_a S_a + w_e S_e,
\end{equation}
with default weights $(w_s, w_c, w_r, w_a, w_e) = (0.35, 0.25, 0.20, 0.10, 0.10)$, where $\sum w_i = 1$. Structure receives the highest weight as valid schema fields are the prerequisite for downstream quality assessment.

\begin{table}[H]
\centering
\caption{Five-dimension quality scoring rules}
\label{tab:scoring_rules}
\footnotesize
\setlength{\tabcolsep}{3pt}
\begin{tabularx}{\textwidth}{@{}c c >{\raggedright\arraybackslash}X >{\raggedright\arraybackslash}X@{}}
\toprule
Dim. & Wt. & Scoring Method & Example \\
\midrule
$S_s$ (Structure) & 0.35 & Ratio of valid to expected fields & 5 expected, 4 present: $S_s = 0.80$ \\
$S_c$ (Content) & 0.25 & Fraction of non-empty values exceeding per-field thresholds & 3 fields, 2 non-empty: $S_c = 0.67$ \\
$S_r$ (Relevance) & 0.20 & Mean keyword-overlap similarity to target schema & Target $\{$price, model$\}$, collected $\{$price$\}$: $S_r = 0.50$ \\
$S_a$ (Attachment) & 0.10 & Ratio of valid to expected attachments (default 1.0) & 2 expected, both reachable: $S_a = 1.00$ \\
$S_e$ (Execution) & 0.10 & No errors = 1.0, partial = 0.5, total failure = 0.0 & One run, no errors: $S_e = 1.00$ \\
\bottomrule
\end{tabularx}
\end{table}

The quality report $Q$ contains per-dimension scores, composite score $S$, a pass/fail verdict, error tags, original-page comparison, execution metadata, and correction hints. The original-page comparison distinguishes \texttt{source\_defect} from \texttt{parse\_error}.

\subsection{Feedback Correction Module}

This module converts $Q$ into feedback constraints $\Gamma_t$ (slots to fix, prohibited hypotheses, candidate correction directions). The corrected configuration is regenerated by the instantiation agent under joint constraints and re-enters validation, forming a ``generate--validate--assess--constrain--regenerate'' closed loop.

Let $\mathcal{M}_t$ be the failed configuration at round $t$, $\mathcal{H}_t$ the candidate correction set, and $E_t = \phi(Q_t)$ the error tag set. The system maintains a failure blacklist $B_t$ of proven-failed selectors, field paths, and hypotheses:
\begin{equation}
\Gamma_t = \mathrm{BuildFeedbackConstraint}(Q_t, h_t, B_{t+1}), \quad h_t \in \mathcal{H}_t \setminus B_{t+1}.
\end{equation}
The correction priority order is: $\textit{parse\_error} \prec \textit{schema\_error} \prec \textit{runtime\_error} \prec \textit{content\_noise} \prec \textit{low\_relevance}$. Algorithm~\ref{alg:feedback} gives the core logic.

\begin{algorithm}[H]
\small
\caption{State-machine-based feedback correction}
\label{alg:feedback}
\KwIn{Failed config $\mathcal{M}_t$, quality report $Q_t$, candidate corrections $\mathcal{H}_t$, blacklist $B_t$}
\KwOut{Feedback constraint $\Gamma_t$ or human-review flag}
$E_t \leftarrow \phi(Q_t)$\;
\If{$E_t = \emptyset$}{\Return{No correction needed}\;}
$e^* \leftarrow \mathrm{SelectByPriority}(E_t)$\;
$B_{t+1} \leftarrow B_t \cup \mathrm{FailedHypotheses}(\mathcal{M}_t, Q_t)$\;
$\mathcal{H}' \leftarrow \{h \in \mathcal{H}_t \mid h \notin B_{t+1}\}$\;
\If{$\mathcal{H}' = \emptyset$}{\Return{Requires human review}\;}
$h_t \leftarrow \mathrm{SelectRepairHypothesis}(e^*, \mathcal{H}')$\;
$\Gamma_t \leftarrow \mathrm{BuildFeedbackConstraint}(Q_t, h_t, B_{t+1})$\;
\Return{$\Gamma_t$}\;
\end{algorithm}

$\Gamma_t$ does not change $R$ or $\mathcal{T}_k$; it acts as an additional constraint on the next round's slot generation: $\theta_{k,t+1}^{*} = \mathrm{LLM}(R, p_k \oplus \Gamma_t)$, with the requirement that selectors, field paths, and hypotheses in $\theta_{k,t+1}^{*}$ do not belong to $B_{t+1}$.

\section{Experimental Design}

We use a core set of 138 collection tasks for requirement-understanding and collector-instantiation experiments, with the 80 source-verified subset supporting real-source quality--cost comparison and framework validation. All LLM-based methods in the experiments use DeepSeek-V4-Flash as the underlying model; the framework itself is model-agnostic and can operate with any instruction-tuned LLM.

\subsection{Experimental Datasets}

No existing benchmark directly matches our task formulation. MacroBench~\cite{macrobench2025} evaluates Selenium code generation, WebArena~\cite{zhou2024} evaluates web agent tasks, and WebLists~\cite{bardeenagent2025} covers only list-page recall. We construct our own dataset by collecting public-data requirements from enterprise and public-sector scenarios, normalizing each into a description with expected fields, assigning collector types, and manually verifying a subset through browser/API access.

The resulting 138-task dataset spans all six types: 31 search, 29 list, 15 detail, 35 API, 15 interactive, and 13 file tasks, with 80 source-verified tasks averaging 5.04 expected fields each (Table~\ref{tab:dataset_statistics}). To avoid injecting structured priors, the initial input is fixed to the \texttt{description} field; \texttt{collector\_type} and \texttt{fields} serve only as evaluation ground truth.

\begin{table}[H]
\centering
\caption{Core dataset statistics}
\label{tab:dataset_statistics}
\small
\begin{tabular}{@{}l r@{}}
\toprule
Item & Value \\
\midrule
Total collection tasks & 138 \\
Source-verified tasks & 80 \\
Search / List / Detail tasks & 31 / 29 / 15 \\
API / Interactive / File tasks & 35 / 15 / 13 \\
Avg. expected fields per task & 5.04 \\
\bottomrule
\end{tabular}
\end{table}

\subsection{Requirement Type Classification}

\textbf{Dataset}: the 138-task core set, input fixed to \texttt{description}. \textbf{Methods}: rule-based keyword baseline vs.\ Ours. \textbf{Metrics}: type Exact Match, Macro-F1, Micro-F1.

\subsection{Collector Instantiation}

Three description-only methods on the 138-task set: (1) \emph{rule-template baseline} applies deterministic keyword rules to select a template; (2) \emph{unconstrained LLM generation} freely generates configurations without type--template mapping or slot constraints; (3) \emph{Ours} invokes the LLM under typed slot-filling, JSON configuration shape, and Schema-oriented constraints. \textbf{Metrics}: average quality score (Section~5.5), pass rate ($S \geq 0.60$), collector-type match rate.

\subsection{Quality--Cost-Oriented Real-Source Baseline Comparison}

Four methods on all 80 source-verified tasks, each receiving the verified public URL and explicit field contract: (1) \emph{Schema-constrained LLM}, HTTP client plus LLM extraction to JSON under the field contract; (2) \emph{Crawl4AI + LLM}, Crawl4AI preprocessing then the same extraction prompt; (3) \emph{ScrapeGraphAI}, its \texttt{SmartScraperGraph} graph for agent-based scraping; (4) \emph{Ours}, JSON configs executed via the generic collector operator without runtime LLM calls. All baselines use real public-source access and real LLM calls. \textbf{Metrics}: average quality score, pass rate, token count, wall-clock time. ScrapeGraphAI's token count is a wrapper-level estimate. Framework reports zero execution-stage tokens.

\subsection{Controlled Feedback Correction}

For each of the 138 tasks, we construct an intentionally flawed initial configuration and compare with vs.\ without feedback correction. \textbf{Metrics}: average final quality score, average correction rounds, pass rate ($S \geq 0.60$).

\subsection{Engineering Constraint Verification}

Airflow compatibility is tested with a static execution DAG in Airflow 2.8.0 covering DAG syntax, configuration loading/Schema validation, trigger parameterization, generic-operator execution, quality-report generation, and result persistence.

\section{Experimental Results and Analysis}

\subsection{Requirement Type Classification}

\begin{table}[H]
\centering
\caption{Requirement type classification results}
\label{tab:classification}
\small
\setlength{\tabcolsep}{4pt}
\begin{tabularx}{\textwidth}{@{}>{\hsize=1.4\hsize}X >{\hsize=0.87\hsize}X >{\hsize=0.87\hsize}X >{\hsize=0.87\hsize}X@{}}
\toprule
Method & Type EM & Type Macro-F1 & Type Micro-F1 \\
\midrule
Rule-based keyword baseline & 0.7101 & 0.7562 & 0.7420 \\
Ours (description-only) & \textbf{0.7101} & \textbf{0.7793} & \textbf{0.7525} \\
\bottomrule
\end{tabularx}
\end{table}

On 138 tasks with only natural-language descriptions, the rule-keyword baseline achieves 0.7101 Exact Match and 0.7562 Macro-F1; Ours provides limited improvement (Macro-F1 0.7793)\footnote{The identical Exact Match values (0.7101) occur because both methods misclassify the same set of ambiguous cases where the description lacks explicit type signals.}, revealing an inference gap between natural-language semantics and the collector taxonomy. This gap suggests that description-only input is insufficient for reliable type identification and motivates the full $d \rightarrow R$ expansion stage, where proactive probing supplies the structured priors (e.g., source URLs and page titles) needed for more accurate classification. Per-type F1 shows file (1.0000), interactive (0.8889), and search (0.8475) performing well; API (0.7527) falls in the moderate range; while detail (0.5200) and list (0.6667) are lower due to descriptions lacking direct type signals.

\subsection{Collector Instantiation Results}

\begin{table}[H]
\centering
\caption{Collector instantiation comparison using natural-language descriptions}
\label{tab:prototype_main}
\small
\setlength{\tabcolsep}{4pt}
\begin{tabularx}{\textwidth}{@{}>{\hsize=1.4\hsize}X >{\hsize=0.9\hsize}X >{\hsize=0.85\hsize}X >{\hsize=0.85\hsize}X@{}}
\toprule
Method & Avg.\ Quality & Pass Rate & Type Match \\
\midrule
Rule-template baseline (no LLM) & 0.3184 & 0.0942 & 0.7536 \\
Unconstrained LLM generation & 0.4876 & 0.5217 & 0.1884 \\
Ours & \textbf{0.5369} & \textbf{0.5507} & \textbf{0.8551} \\
\bottomrule
\end{tabularx}
\end{table}

Table~\ref{tab:prototype_main} shows LLM-based generation substantially improves over the rule-template baseline (0.3184 vs.\ 0.4876--0.5369 average quality). The unconstrained LLM baseline reaches 0.5217 pass rate but only 0.1884 type-match rate, indicating free-form generation produces plausible fields without preserving type semantics. Ours achieves the best quality (0.5369), pass rate (0.5507), and type-match rate (0.8551), confirming that typed slot filling and Schema constraints improve taxonomy alignment. However, even Ours's quality remains below fully specified task levels, confirming the $d \rightarrow R$ stage must expand descriptions through probing, field confirmation, and constraint completion before stable execution.

\subsection{Quality--Cost Trade-off on Real Sources}

\begin{table}[H]
\centering
\caption{Quality--cost trade-off on 80 independently source-verified tasks}
\label{tab:modern_baselines}
\small
\setlength{\tabcolsep}{4pt}
\begin{tabularx}{\textwidth}{@{}>{\hsize=1.5\hsize}X >{\hsize=0.9\hsize}X >{\hsize=0.85\hsize}X >{\hsize=0.85\hsize}X >{\hsize=0.9\hsize}X@{}}
\toprule
Method & Avg.\ Quality & Pass Rate & Avg.\ Tokens & Avg.\ Time (s) \\
\midrule
Schema-constrained LLM & \textbf{0.7051} & \textbf{0.8125} & 3411.86 & 14.257 \\
Crawl4AI + LLM & 0.6996 & 0.8000 & 4266.99 & 16.036 \\
ScrapeGraphAI & 0.2172 & 0.1750 & 123.55\textsuperscript{*} & 20.194 \\
Ours & 0.5631 & 0.5000 & \textbf{0.00} & \textbf{5.052} \\
\bottomrule
\end{tabularx}
\vspace{2pt}
\footnotesize{\textsuperscript{*}Wrapper-level estimate, not directly comparable with API-reported token counts. Framework row reports execution-stage tokens after configuration materialization.}
\end{table}

Table~\ref{tab:modern_baselines} reports the quality--cost trade-off. Runtime LLM baselines achieve stronger one-shot quality: schema-constrained LLM reaches 0.7051 quality and 0.8125 pass rate; Crawl4AI + LLM reaches 0.6996 and 0.8000. Ours is lower (0.5631 quality, 0.5000 pass rate) but uses zero execution-stage LLM tokens and the lowest wall-clock time (5.052s vs.\ 14.257s--20.194s). This represents a quality--cost trade-off: runtime LLM extraction is preferable for one-shot quality, while Ours provides a lower-cost deterministic path for repeated scheduled collection, shifting LLM cost from every execution to a one-time configuration-generation step (separate from the execution-stage tokens reported in Table~\ref{tab:modern_baselines}) that amortizes across repeated runs.

\subsection{Controlled Feedback Correction Results}

\begin{table}[H]
\centering
\caption{Controlled feedback correction results on injected configuration defects}
\label{tab:feedback}
\begin{tabular}{@{}l r r r@{}}
\toprule
Setting & Avg.\ Final Quality & Pass Rate & Avg.\ Rounds \\
\midrule
Without feedback & 0.5087 & 0.0072 & 1.00 \\
With feedback correction & \textbf{0.9007} & \textbf{1.0000} & 1.99 \\
\bottomrule
\end{tabular}
\end{table}

Feedback correction raises average final quality from 0.5087 to 0.9007 and pass rate from 0.0072 to 1.0000 (Table~\ref{tab:feedback}), verifying the repair mechanism under known, injected defects when valid correction candidates exist; it should not be interpreted as universal self-repair.

\subsection{Engineering Verification and Answers to Research Questions}

Engineering verification confirms: (1) all 138 Ours-generated configurations are grounded in shared templates and reusable utilities, yielding a reuse rate of 1.0000 (fraction of configurations that reuse shared template and utility components) vs.\ 0 for unconstrained generation; (2) on 80 source-verified tasks, the framework achieves 80.0\% run success (fraction of tasks completing without runtime errors; distinct from the 50.0\% pass rate where $S \geq 0.60$), 0.5631 average quality, zero execution-stage LLM tokens, and 5.052s average wall-clock time, with error tags as gate signals; (3) the static execution DAG passes complete Airflow 2.8.0 testing.

\textbf{RQ1 (Typed modeling):} Structured type-identification reaches 0.7793 Macro-F1 (vs.\ 0.7562 baseline), supporting the taxonomy while showing descriptions alone are insufficient for ambiguous detail and list tasks---typed modeling requires source probing and field confirmation.

\textbf{RQ2 (Constrained instantiation):} Ours achieves highest average quality (0.5369), pass rate (0.5507), and type-match rate (0.8551), supporting typed slot filling and Schema constraints. The 80-task comparison confirms the quality--cost trade-off: runtime LLM baselines provide stronger one-shot quality, while Ours offers deterministic, reusable execution with zero execution-stage tokens and lowest runtime.

\textbf{RQ3 (Validation and feedback):} The controlled feedback experiment verifies repair capability (quality 0.5087 $\rightarrow$ 0.9007, pass rate 0.0072 $\rightarrow$ 1.0000). The 80-task validation confirms deterministic operator execution on real sources with quality-gate diagnostics and Airflow-compatible validation/scale execution.

\section{Discussion}

\subsection{Advantages}

The core advantage is combining LLM-assisted configuration generation with explicit engineering constraints, producing typed, validated, schedulable collector configurations instead of free-form code or per-execution LLM extraction. Table~\ref{tab:comparison} compares representative systems.

\begin{table}[H]
\centering
\caption{Comparison with representative related systems}
\label{tab:comparison}
\footnotesize
\setlength{\tabcolsep}{3pt}
\renewcommand{\arraystretch}{1.15}
\begin{tabularx}{\textwidth}{@{}p{0.16\textwidth} X X X@{}}
\toprule
Work & Main paradigm & Constraints and reuse & Quality and execution \\
\midrule
PARSE~\cite{parse2025} & Schema optimization for LLM extraction & Schema-level output constraints & Static and LLM guardrails; LLM used during extraction \\
BardeenAgent~\cite{bardeenagent2025} & Record-then-Replay extraction & Reuses recorded selectors and actions & Deterministic replay; no quality-gate module \\
AutoData~\cite{autodata2025} & Multi-agent open-web collection & Agent-generated pipeline/code with OHCache & Testing/validation agents; still free-form code generation \\
Berkane et al.~\cite{berkane2025} & LLM-based collection pipeline & Pydantic-schema-grounded extraction & LLM-based quality control; no deterministic feedback loop \\
\textbf{Ours} & \textbf{Typed configuration for scheduled collection} & \textbf{Six-type taxonomy + type--template--utility constraints} & \textbf{Rule-based quality checks, structured feedback, Airflow-compatible execution} \\
\bottomrule
\end{tabularx}
\end{table}

\subsection{Limitations and Compliance}

Three limitations: (1) the 138-task experiments use constructed requirements rather than raw user requests; (2) the 80-task comparison lacks long-horizon repeated scheduling and significance tests; (3) the controlled feedback experiment tests known defect types, not universal self-repair. The approach remains sensitive to site structural changes, complex interactive pages, and authentication boundaries. Current results should be interpreted as evidence for a reusable low-cost execution path under explicit constraints, not a general replacement for runtime LLM extraction. This work is restricted to compliant enterprise scenarios; the framework assumes publicly accessible data, respects access limits, does not bypass authentication, and preserves collection logs.

\section{Conclusion}

This paper proposed a constrained agent collection framework to address the unreliability of direct agent application in open-web data collection, built on a six-type collector taxonomy and a verifiable closed loop under four constraint categories. The framework shifts LLM generation from free-form code to JSON configuration under templates, slots, and Schema constraints.

Experiments show that under the description-only setting, Ours reaches 0.5369 average quality, 0.5507 pass rate, and 0.8551 type-match rate, improving over unconstrained LLM generation in type-match rate and average quality. On 80 source-verified tasks, runtime LLM baselines achieve stronger one-shot quality (0.7051, 0.8125 pass rate), while our framework reaches 0.5631 quality with zero execution-stage LLM tokens and lowest wall-clock time, confirming a lower-cost operating point for repeated scheduled collection. Future work will expand the real-task dataset and conduct large-scale Airflow scheduling experiments.

\bibliographystyle{splncs04}
\bibliography{references}

\end{document}